\documentclass[sigconf, nonacm]{acmart}
\makeatletter
\def\@ACM@checkaffil{
    \if@ACM@instpresent\else
    \ClassWarningNoLine{\@classname}{No institution present for an affiliation}%
    \fi
    \if@ACM@citypresent\else
    \ClassWarningNoLine{\@classname}{No city present for an affiliation}%
    \fi
    \if@ACM@countrypresent\else
        \ClassWarningNoLine{\@classname}{No country present for an affiliation}%
    \fi
}
\makeatother
\usepackage{url}
\usepackage[ruled,vlined]{algorithm2e}
\usepackage{pgfplots}
\usepackage{multirow}
\usepackage{multicol} 
\usepackage{dcolumn}
\usepackage{graphicx}
\usepackage{subcaption}
\usepackage{hyperref}
\usepackage{float}
\usepackage{algpseudocode}
\usepackage{amsmath}
\usepackage{balance}
\usepackage[a-1b]{pdfx}
\usepackage{hyperref}
\usepackage{graphicx}
\usepackage{listings}
\usepackage{pythonhighlight}
\usepackage{listings}
\usepackage{hyperref}
\usepackage{url}

\hypersetup{
    colorlinks=true,
    linkcolor=blue,
    filecolor=magenta,      
    urlcolor=cyan,
    pdftitle={Overleaf Example},
    pdfpagemode=FullScreen,
    }
\usepackage{xcolor}

\lstdefinelanguage{JSON}{
    morestring=[b]",
    morestring=[s]{""}{"},
    morecomment=[l]{//},
    morecomment=[s]{/*}{*/},
    morekeywords={true, false, null},
    sensitive,
    basicstyle=\ttfamily,
    commentstyle=\color{green},
    stringstyle=\color{blue},
    keywordstyle=\color{blue},
    showstringspaces=false,
    breaklines=true,
}

\definecolor{codegreen}{rgb}{0,0.6,0}
\definecolor{codegray}{rgb}{0.5,0.5,0.5}
\definecolor{codepurple}{rgb}{0.58,0,0.82}
\definecolor{backcolour}{rgb}{1.0,1.0,1.0}

\begin{document}
\title{Minimum Tuning to Unlock Long Output from LLMs with High Quality Data as the Key}
\author{Yingda Chen$^{\dagger \star}$, Xingjun Wang$^{\dagger}$, Jintao Huang, Yunlin Mao, Daoze Zhang and Yuze Zhao}\thanks{$\dagger$: equal contributions. \\ $\star$: corresponding author.}
\affiliation{%
  \institution{ModelScope Team, Alibaba Group \\ 
  }
}

\begin{abstract}
As large language models rapidly evolve to support longer context, there is a notable disparity in their capability to generate output at greater lengths. Recent study\cite{bai2024longwriterunleashing10000word} suggests that the primary cause for this imbalance may arise from the lack of data with long-output during alignment training. In light of this observation, attempts are made to re-align foundation models with data that fills the gap, which result in models capable of generating lengthy output when instructed. In this paper, we explore the impact of data-quality in tuning a model for long output, and the possibility of doing so from the starting points of human-aligned (instruct or chat) models.  With careful data curation, we show that it possible to achieve similar performance improvement in our tuned models, with  only \textit{a small fraction} ($3.74\%$) of training data instances and compute. In addition,  we assess the  \textit{generalizability} of such approaches by applying our tuning-recipes to several models. our findings suggest that, while capacities for generating long output vary across different models  out-of-the-box, our approach to tune them with high-quality data using lite compute, consistently yields notable improvement across all models we experimented on. We have made public our curated dataset for tuning long-writing capability, the implementations of model tuning and evaluation, as well as the fine-tuned models, all of which can be openly-accessed.
\end{abstract}

\maketitle

\section{Introduction}
Despite recent strides in large language models (LLMs) for handling extensive inputs\cite{Anthropic-sonnet,reid2024gemini}, current long-context LLMs face a notable challenge when it comes to generating equally lengthy outputs. The ability to produce coherent and contextually rich text over extended lengths remains a critical bottleneck. This limitation not only restricts the practical utility of these models but also highlights a gap between input processing and output generation capabilities. As the demand for more sophisticated and versatile AI systems grows, addressing this disparity becomes increasingly important.

The capability to handling long context does not readily translate into that of outputting long output following the prompted instruction. To tackle the imbalance of these two capabilities normally found in general-purposed LLMs, several previous studies have gone into instructing the LLM to follow length constraints\cite{yuan2024followinglengthconstraintsinstructions}, and to understand multi-constraint instruction for long-form text generation\cite{pham2024surimulticonstraintinstructionfollowing}. In particular, the authors of \cite{bai2024longwriterunleashing10000word} make the important observations that for many LLMs, there is a significant lack of alignment-training data samples with lengthy output. In recognition of this gap, an alignment-based approach was taken to align base model with data of lengthy output, resulting in LongWriter bilingual model\cite{longwriter-model} that scales the output window size of  LLMs to $10,000+$ words. Building upon of these earlier observations, we are of further opinion that not only the length of output matters, the proper match between the required length in the prompt instruction, and the actual output length, also matters in tuning data. Therefore, it is important to tune model with high-quality dataset that meet the requirements not only in length, but also in coherence. In this paper, we show how to curate high-quality dataset that meets both of these requirements, and tune model for the task of long-form length-following. In addition, with high-quality data, it is possible to achieve the goal with minimum tuning cost, as measured by the number of training data instances. By doing so, we aim to provide insights into how to tune models that handle both long inputs and generate high-quality, extended outputs, and at a much lower cost.

Specifically, we have made the following contributions in this paper:
\begin{itemize}
\item \textbf{Identification of the Importance of Aligning Quality of Data with the Task:} We identify that aligning the quality of the data is crucial for tuning model towards long-writing task. Once high-quality data that align well with the task, it is possible to achieve significant model performance improvement with as little as $666$ data samples.
\item \textbf{Human-Aligned Model as a Starting Point:} Our results confirm that a well trained, general human-aligned model can serve as a good starting point for unlocking the capability of outputting long text and following prompted instruction. While our work leverages upon, and is motivated by results outlined in \cite{bai2024longwriterunleashing10000word}, we deviate from the alignment-based approach therein, and choose to tune our model from human-aligned starting point. We believe this is a viable solution that should apply to a wide range of tasks that do not conflict with the overall goal of human-alignment. 
\item \textbf{Efficient Tuning with Minimal Data:} Altogether, we show that by starting tuning from human-aligned models, and by using a small quantity of high-quality data, a much higher tuning efficiency can be achieved. In particular, we are able to tune a long-writing model with similar performance\cite{bai2024longwriterunleashing10000word} with only less than $\mathbf{4\%}$ of training data instances.
\end{itemize}

The rest of this paper is organized as follows. Section \ref{sec:evaluation-methodologies} outlines the methodologies for evaluating a model's performance for the task of long-form output following instructions. The process of data curation to produce high-quality data is then described in Section \ref{subsec:data-curation}. We than present how we tune the models for the target tasks with different experiments, and present the performance results of the models tuned under various setups. The paper is then concluded in Section \ref{sec:conclusion}, with additional data presented in Appendix.

\section{Evaluation Methodologies}
\label{sec:evaluation-methodologies}
We followed the evaluation methodologies in \cite{bai2024longwriterunleashing10000word} and adopted \texttt{LongBench-Write} as the benchmark for evaluating a model's capacities in following instruction to output long text. In particular, we have integrated an open-source implementation of \texttt{LongBench-Write} in EvalScope framework\cite{longbench-write-evalscope}. With \texttt{LongBench-Write} ,  two metrics are employed to evaluate the quality of a model, namely
\begin{itemize}
    \item Output Length Score $S_L$, and
    \item Output Quality  Score $S_Q$

\end{itemize}
In particular, to ensure that the model's output length is as close as possible to the requirement specified in the instructions, a piece-wise linear function is used to calculate $S_L$: 
\begin{equation}
\label{eqn:sl}
    S_{L}=\left\{\begin{array}{ll}{100 \cdot \max (0, 1-\left(\frac{L^{\prime}}{L}-1\right) / 3)} & {\text { if } L^{\prime}>L,} \\ {100 \cdot \max (0,1-\left(\frac{L}{L^{\prime}}-1\right) / 2)} & {\text { if } L^{\prime} \leq L.}\end{array}\right.
\end{equation}
$L$ denotes the target length as instructed and $L^{\prime}$ is the actual output length. With this metric, an output length falling short or excessively over-length with both lead to lower score of $S_L$. On the other hand, the output quality $S_Q$ is comprehensively evaluated by \texttt{GPT4o} from various perspectives including Relevance, Accuracy, Coherence, Clarity,Breadth and Depth and Reading Experience. Detailed description for these evaluation metric can be found in \cite{bai2024longwriterunleashing10000word} or in our reference implementation\cite{longbench-write-evalscope}.

\section{Data Curation}
\label{subsec:data-curation}
The authors in \cite{bai2024longwriterunleashing10000word} explore the  hypothesis for why most models cannot generate extremely long text outputs, and believe that limitation on output length primarily stems from the characteristics of the supervised fine-tuning (SFT) datasets. Specifically, the maximum generation length of a model is effectively constrained by the upper limit of output lengths presented in its SFT dataset. Therefore, a model's ability to output text of extreme length is impeded since it rarely sees data samples of lengthy output during alignment, and is therefore incapable of doing so. For example, as suggested in \cite{glm2024chatglm,bai2024longwriterunleashing10000word}, long output data is extremely under-represented during alignment, as data with output exceeding $2000$ words only constitute a mere entire $0.1\%$ of entire chat-SFT data. In response to this observation, longer output data is synthesized using AgentWriter to obtain data that conforms to the requirement of lengthy output\cite{bai2024longwriterunleashing10000word}, which is made publicly available\cite{longwriter-6k-ms,longwriter-6k-hf}. The dataset contains a entries with output that span evenly across ranges of $[0,500]$, $[500,2000)$,$[2000,4000)$, $[4000,2000+)$. In conjunction with the GLM's SFT data, these data was used to train the LongWriter model \cite{longwriter-model}.

\subsection{Inspection of LongWriter-6K Dataset}
\label{subsec:6k-data-quality}
Despite efforts to make  \texttt{LongWriter-6K} dataset suitable for training model to output long-form text, upon careful review, we have found that there is still room for improvement in \texttt{LongWriter-6K}. In particular, we identify that substantial amount of sub-optimal data samples exist in the dataset that may not align well with the task. By inspecting the correlation between data characteristics and the evaluation methodologies outlined in Section \ref{sec:evaluation-methodologies},  we conclude that, to align data better with the task of tuning a model to output lengthy text of given length, different categories of sub-optimal data entries shall be cleansed, including:
\begin{itemize}
    \item \textbf{Category 1:}  Data samples  with input prompt that does \textit{not} specify ``Required Length'' for output. 
        \begin{lstlisting}[language=JSON]
{
  "messages": [
    {"role": "user", "content": "Can you please write a  Stakeholder Analysis and Engagement Strategy for Online Loan Tracking System software project?\n"}, 
    {"role": "assistant", "content": "### Introduction to the Online Loan Tracking System Software Project ..."}]
}
        \end{lstlisting}
    One of such example is shown above. Such sub-optimal entry would \textit{unlikely} guide the model to align with the target task, as no length requirement is specified, and there is no way for model-training to learn expected behavior from such data. Upon review of the dataset, we found a significant proportion ($70.9\%$) of data in \texttt{LongWriter-6K} fall into this category.
    
    \item \textbf{Category 2:} Data samples that exhibit significant discrepancy between the ``Required Length'' in prompt, and the actual ``Output Length''. 
\begin{lstlisting}[language=json]
{"messages": [{"role": "user", "content": "Write in 50000 words a romanace, fantasy, mystery, adventure, sexy and drama bl love story about vampires and werewolves place and time specified it must have some comedy scenes too and a happy ending. "}, {"role": "assistant", "content": "### The Enigmatic Night\n\nIn the heart of a sprawling metropolis, where neon lights reflected off rain-slicked streets and the city pulsed with an electric, nocturnal energy, lived a young vampire named Alex ..."}], "length": 50000, "response_length": 16199}
        \end{lstlisting}
    One of such example is shown above, where a ``Required Length'' of $50,000$ is specified in input instruction while the output measured at only $16,199$. In addition, we found that a common discrepancy between ``Required Length'' and  ``Output Length'' arise when a data sample contains response of \textit{desk-rejection} from a model. In this case, the output resembles response along the line of ``\textit{I cannot answer this question for safety reasons, please try other questions.}''. One can imagine that, such entries in dataset would likely confuse the model alignment, or even pollute the model's capability to follow instruction.
\end{itemize}

\begin{figure}[!htp]
\includegraphics[width=0.5\textwidth]{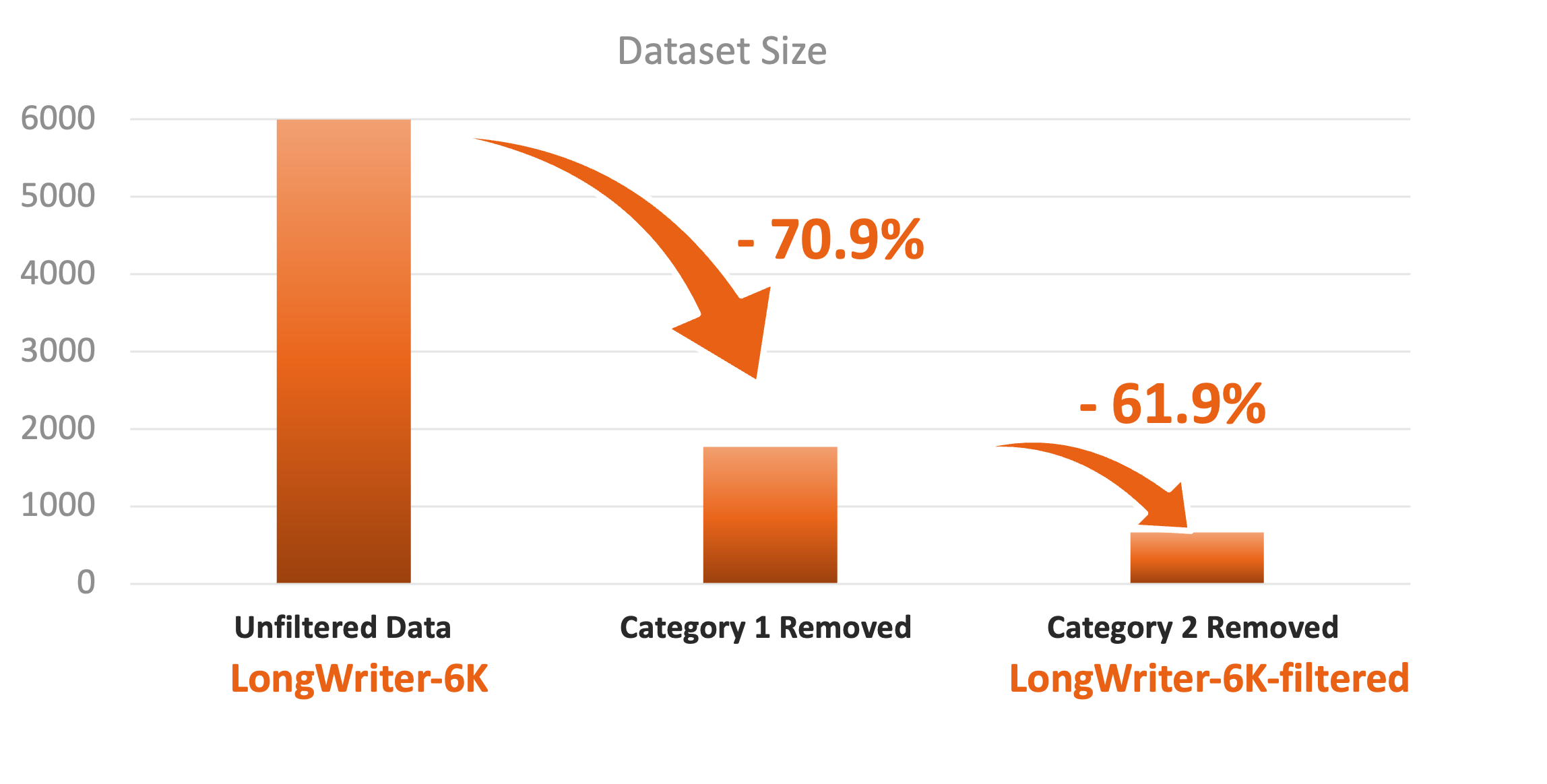}
  \caption{\textmd{Two-Stage Data Refinement on \texttt{LongWriter-6K}}}
  \label{fig:dataremoval}
\end{figure}
\subsection{Data Refinement}
\label{subsec:data-refinement}
As has been shown in earlier studies \cite{zhou2024lima}, \textit{quality} of the alignment data can be of extreme importance to model's performance, more so than the \textit{quantity} of the data. It is therefore imperative to further improve the quality to better align with the target task.

In light of the two categories of data sub-optimalities, we take  \texttt{LongWriter-6K} as the starting dataset and perform further two-step ``data cleansing''. First we remove all data entries missing ``Required Length'' in input instruction, reducing the total $6000$ data entries to $1748$. Additionally we strip away data that display noticeable discrepancy between ``Required Length'' and  ``Output Length''. More specifically, we use \ref{eqn:sl} as the eliminating metric, and remove all entries with a score of $S_L < 80$. Doing so further reduce the count of data entries from $1748$ to $666$: only about $10\%$ of the \texttt{LongWriter-6K} remain after the two-step filtration. The process is illustrated in Fig. \ref{fig:dataremoval}, which result in a new dataset \texttt{LongWriter-6K-filtered}, with $666$ high-quality data entries that align well with the task of ``generating output at required length''. 

\begin{figure}[hbp] 
    \centering
    \begin{subfigure}[b]{0.23\textwidth}  
        \includegraphics[width=\textwidth]{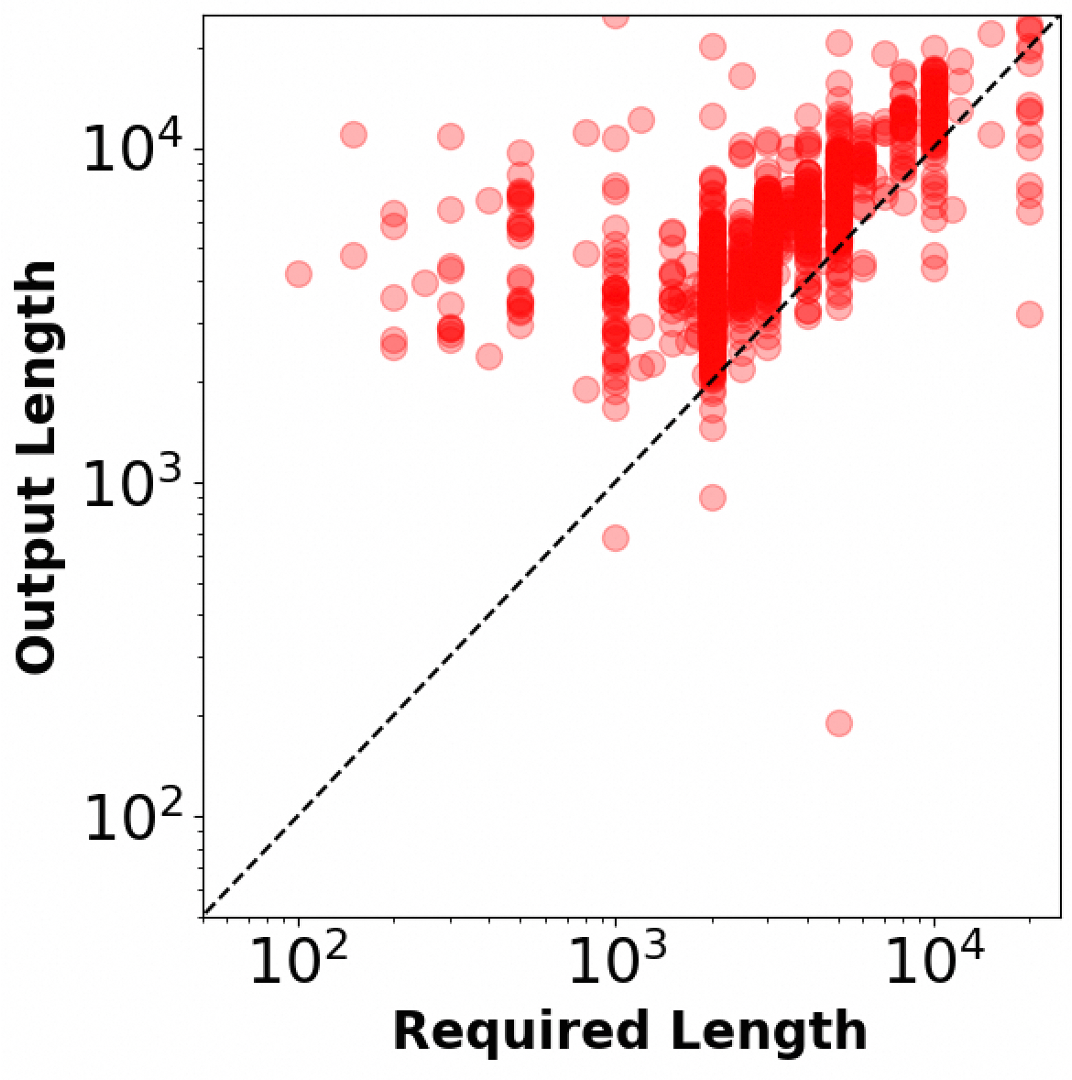}  
        \caption{Intermediate dataset: after removing Category 1 entries.}
        \label{fig:data-characteritics-sub1}
    \end{subfigure}
     \hfill  
    \begin{subfigure}[b]{0.23\textwidth}  
        \includegraphics[width=\textwidth]{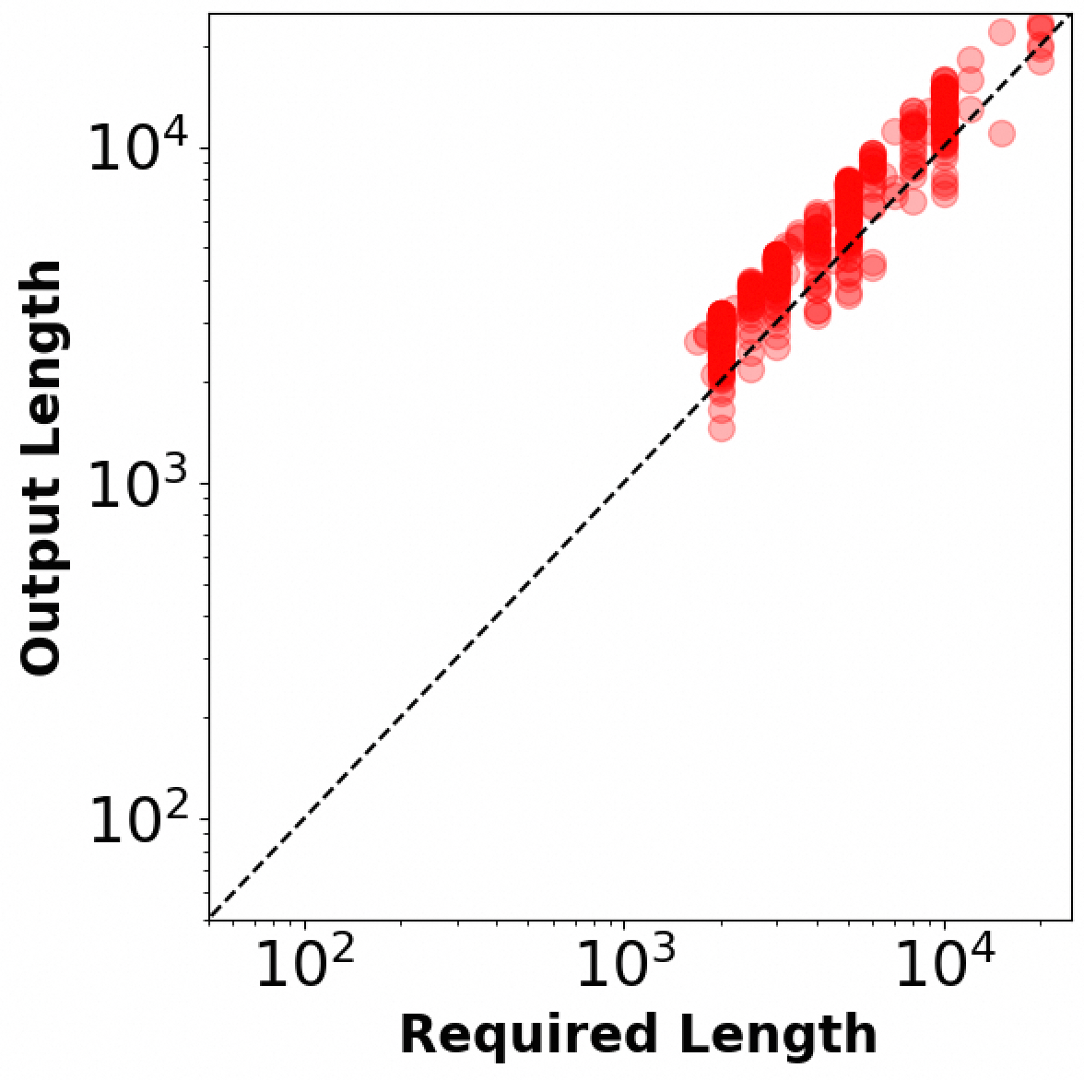}
        \caption{\texttt{LongWriter-6K-filtered}: after removing Category 2 entries. }
        \label{fig:data-characteritics-sub2}
    \end{subfigure}
    \caption{\textmd{Comparison of different dataset's ``length-following'' characteristics.}}
    \label{fig:data-characteritics}
\end{figure}

Fig. \ref{fig:data-characteritics} depicts data characteristics after applying the filtration of Category 1 and 2 data. As we can see, after two rounds of filtering, the actual ``Output Length'' of the data now demonstrates a very good adherence to the ``Required Length'', with an overall near-linear alignment between the lengths of the data samples. The resulting \texttt{LongWriter-6K-filtered} dataset containing 666 entries, which is made publicly available \cite{longwriter-6k-filtered}, and the data processing implementation is also made open-sourced\cite{longbench-write-evalscope}.

\begin{table*}[!htp]
\label{tab:qwen2-result}
\centering
\begin{tabular}{c|l|c c c|c}
\hline
 & \centering{\textbf{Training data and epoch setup}} & \textbf{S (avg)} & \textbf{$S_L$} & \textbf{$S_Q$} & \textbf{Improvement} \\ \hline
\texttt{Qwen2-7B-Instruct}  & -- & 67.4 & 48.92 & 85.87 & -- \\ \hline
\multirow{2}{*}{Setup 1} & LongWriter-6k with 1:1 mixing  ratio& \multirow{2}{*}{66.38} & \multirow{2}{*}{52.76} & \multirow{2}{*}{80.00} & \multirow{2}{*}{-1.02 pt} \\
 &  of alignment data(6k), trained for 2 epochs &  &  &  &  \\ \hline
\multirow{2}{*}{Setup 2} & LongWriter-6k with 1:30 mixing ratio & \multirow{2}{*}{72.02} & \multirow{2}{*}{64.59} & \multirow{2}{*}{79.44} & \multirow{2}{*}{+4.62 pt} \\
 & of alignment data(180k), trained for 1 epoch &  &  &  &  \\ \hline
\multirow{2}{*}{Setup 3} & \textbf{LongWriter-6k-filtered} only, & \multirow{2}{*}{76.62} & \multirow{2}{*}{70.44} & \multirow{2}{*}{82.8} & \multirow{2}{*}{+9.22 pt} \\
 & trained for 4 epochs &  &  &  &  \\ \hline
\multirow{2}{*}{\textbf{Setup 4}} & \textbf{LongWriter-6k-filtered} with 1:20 mixing ratio  & \multirow{2}{*}{79.51} & \multirow{2}{*}{75.61} & \multirow{2}{*}{83.4} & \multirow{2}{*}{+12.1 pt} \\
 & of alignment data(13k), trained for 2 epochs &  &  &  &  \\ \hline
\end{tabular}
\caption{Tuning \texttt{Qwen2-7B-Instruct} with various setups for long-writing task.}
\label{tab:qwen2-result}
\end{table*}

\section{Tuning With Curated Data}
It is now recognized that given a well-trained foundation model, the quality of SFT data is pivotal in determining the performance and capacity of the models after alignment\cite{zhou2024lima}.  In this section, we explore the task of tuning a model to generate long output. While our work is built upon, and has taken inspirations from \cite{bai2024longwriterunleashing10000word}, our approach is fundamentally different in several aspects:
\begin{itemize}
    \item First of all, we believe that for the task of \textit{generating long text following instructed length}, it may not be necessary to build from a un-aligned base model and train with the entire SFT phase. Instead, with proper tuning techniques, starting off from a human-aligned \textit{instruct} or chat version is a plausible approach. Therefore, unlike the approach in \cite{bai2024longwriterunleashing10000word}, we have chosen to tune instructed and chat models instead. After all, the capability of writing long output is more complementary to,  than being in conflict with, the human-aligned capability of a model.
    \item Secondly, as detailed in Sec. \ref{subsec:data-refinement}, we perform thorough data cleansing, and after the two-stage distillation, the curated \texttt{LongWriter-6k-filtered} dataset  represents a collection of data that aligns well with the target task, which we believe to be  pivotal in shaping the behavior of fine-tuned models. 
\end{itemize}
The different approaches we adopt resulted in a large amount of saving in tuning cost, which we discuss in this Section.

\subsection{Model Selection}
To validate the general applicability of the our approach, which is extended from the original LongWriter\cite{bai2024longwriterunleashing10000word} training strategy based on GLM model families,  we have chosen \texttt{Qwen2-7B-Instruct}\cite{yang2024qwen2} and \texttt{GLM4-9B-Chat}\cite{glm2024chatglm} as the starting baseline for model-tuning. The reason for selecting human-aligned models have been discussed above. In addition, we added the latest \texttt{Qwen2.5-7B-Instruct}\cite{qwen2.5} just released as we were finishing up our study. As the latest addition to the Qwen model families, Qwen-2.5 series improve on the capacity on ``generating long texts'', and provide meaningful comparison with its predecessor Qwen-2 model.

\subsection{Datasets for Model Tuning}

The datasets used in different experiments are listed below:
\begin{itemize}
    \item \texttt{LongWriter-6k}: $6000$ data entries
    \item \texttt{LongWriter-6k-filtered}: $666$ data entries
    \item \texttt{MagPie-Qwen2-Pro-200K-English}\cite{magpie-qwen2-english}: $200,000$ data entries
    \item \texttt{MagPie-Qwen2-Pro-200K-Chinese}\cite{magpie-qwen2-chinese}: $200,000$ data entries
\end{itemize}
Among this list, \texttt{LongWriter-6k} and \texttt{LongWriter-6k-filtered} are the \textit{task datasets} designed for our task of tuning model for to output long text, while dataset \texttt{MagPie-Qwen2-Pro-200K-English} and \texttt{MagPie-Qwen2-Pro-200K-Chinese} are \textit{alignment data} synthesized using MagPie\cite{xu2024magpie}\footnote{It should be noted that we rely on these two open dataset synthesized with \texttt{Qwen2-72B-Instruct} using MagPie, which may  be sub-optimal for non-Qwen2 models.}. The alignment dataset is introduced to avoid potential degradation in model's generalization ability and catastrophic forgetting\cite{huang2024catastrophic-forgetting-mitigating} with rehearsal methods. In particular, alignment data is sampled at different rates which result in various mixing ratios with task-data in tuning experiments, which we explain in detail later in Section \ref{subsec:experiment}. It is important to note that, for all our experiments, the alignment dataset was only meant to serve as guided data to avoid loss of general capabilities during tuning. Since we have started from human-aligned versions of models, only \textit{a small fraction($3\%$ at the maximum) is sampled }from the alignment dataset and used during turning.


\subsection{Experiment Setup}
All models are tuned on a single node with 4xA100 80G GPUs. We use ModelScope Swift(MS-Swift)\cite{swift-arxiv} for all model tuning, an open-sourced training framework that adapts to various tuning techniques via standardized interfaces. Some of the example tuning scripts are included in Appendix \ref{subsec:codes}. 

To mitigate the reduction in the contribution of target tokens to the overall loss for longer output sequences, which can result in sub-optimal model performance on tasks requiring extended outputs, a token-averaged loss weighting strategy is employed. With this approach, the loss is calculated as the mean of individual losses associated with each target token within a given batch. We implement this particular loss as \texttt{loss-ce}(cross-entropy) in the open-sourced turning framework MS-Swift\cite{swift-arxiv}.

During model evaluation, repetition\_penalty of $1.1$, and temperature of $0.5$ is used for inference. We follow the evaluation mythologies outlined in Section \ref{sec:evaluation-methodologies}. In particular, the evaluation procedures and associated metrics are implemented based on EvalScope\cite{evalscope}. We include exampled evaluation implementation in Appendix \ref{subsec:evaluationscript}. 

\begin{table*}[!htp]
\centering
\begin{tabular}{c|l|c c c|c}
\hline
 & \centering{\textbf{Training data and epoch setup}} & \textbf{S (avg)} & \textbf{$S_L$} & \textbf{$S_Q$} & \textbf{Improvement} \\ \hline
\texttt{GLM4-9b-Chat}  & -- & 67.8 & 52.8 & 82.78 & -- \\ \hline
\multirow{2}{*}{LongWriter-GLM4-9B\cite{bai2024longwriterunleashing10000word,longwriter-model}} & Trained from GLM-4-9B\cite{glm4-model} with LongWrite-6k plus 1:30 mixing ratio    & \multirow{2}{*}{\textbf{80.5}} & \multirow{2}{*}{\textbf{78.6}} & \multirow{2}{*}{82.3} & \multirow{2}{*}{\textbf{+12.7 pt}} \\
 &  using entire GLM chat SFT dataset (180k), trained for 4 epochs &  &  &  &  \\ \hline 
\multirow{2}{*}{\textbf{Setup 4}} & LongWriter-6k-filtered with 1:20 mixing ratio of alignment data (13k),    & \multirow{2}{*}{79.88} & \multirow{2}{*}{77.42} & \multirow{2}{*}{\textbf{82.33}} & \multirow{2}{*}{+12.08 pt} \\
 & trained for 2 epochs &  &  &  &  \\ \hline
\end{tabular}
\caption{Tuning \texttt{GLM4-9b-Chat} with various setups for long-writing task.}
\label{tab:glm4-result}
\end{table*}

\begin{table*}[!htp]
\centering
\begin{tabular}{c|l|c c c|c}
\hline
& \centering{\textbf{Training data and epoch setup}} & \textbf{S (avg)} & \textbf{$S_L$} & \textbf{$S_Q$} & \textbf{Improvement} \\ \hline
\texttt{Qwen2.5-7B-Instruct}  & -- & 75.79 & 66.4 & 85.17 & -- \\ \hline
\multirow{2}{*}{\textbf{Setup 4}} & LongWriter-6k-filtered with 1:20 mixing ratio of alignment data (13k),    & \multirow{2}{*}{79.88} & \multirow{2}{*}{77.42} & \multirow{2}{*}{82.33} & \multirow{2}{*}{+4.75 pt} \\
 & trained for 2 epochs &  &  &  &  \\ \hline
 \multirow{2}{*}{Setup 5} & Based on Setup 4, introduce additional annealing with learning rate    & \multirow{2}{*}{\textbf{82.84}} & \multirow{2}{*}{\textbf{81.24}} & \multirow{2}{*}{\textbf{84.44}} & \multirow{2}{*}{+7.05 pt} \\
 & of $2e-6$ using LongWriter-6k-filtered for another 2 epochs &  &  &  &  \\ \hline
\end{tabular}
\caption{Tuning \texttt{Qwen2.5-7B-Instruct} with various setups for long-writing task.}
\label{tab:qwen2.5-result}
\end{table*}

\subsection{Experiment Results}
\label{subsec:experiment}
We begin by tuning \texttt{Qwen2-7B-Instruct}, by adopting various combinations of the task and alignment datasets. The performance for different tuned models are listed in Table \ref{tab:qwen2-result}. From the observation above, it is evident that tuning with high quality data is crucial for model to achieve superior performance in long-form writing tasks. Most notably:
\begin{itemize}
    \item Tuning with the unfiltered \texttt{LongWriter-6k} dataset from \texttt{Qwen2-7B-Instruct} does not result in significant performance improvement. In particular, while long-output length-following capability improves, noticeable regression in writing quality is introduced. This may be attributed to data sub-optimality analyzed in Section \ref{subsec:6k-data-quality}.
    \item Tuning with 2,664 high-quality data instances (trained with $666$-sample \texttt{LongWriter-6k-filtered}  dataset for $4$ epochs) resulted in in a model  significantly outperforming the model trained on a mixture of \texttt{LongWriter-6k} and 180,000 alignment data samples, in terms of both length-following ($S_L$) and writing quality $S_Q$.     
    \item Proper mixing of task data and alignment data helps not only sustaining the writing quality $S_Q$, by also help improve the length-following capability $S_L$. For example, extending Setup 4 with alignment data sampled from \cite{magpie-qwen2-chinese,magpie-qwen2-english} at $1:20$ mixing ratio ($13320$ data samples coming equally from \cite{magpie-qwen2-chinese} and \cite{magpie-qwen2-english}) achieve the best performance among all Setups, with 2 epochs of tuning.
\end{itemize}
We have made the model checkpoints achieved from \texttt{Setup} 4 public available at name it  \texttt{MS-LongWriter-Qwen2-7B-Instruct} \cite{MS-LongWriter-Qwen2-7B-instruct}. 
This result also validates the generalizbility of tuning models for long-writing tasks, which is not limited to the GLM series of models studied in \cite{bai2024longwriterunleashing10000word}. 

\begin{figure}[!htp] 
    \centering
    \begin{subfigure}[b]{0.23\textwidth}  
        \includegraphics[width=\textwidth]{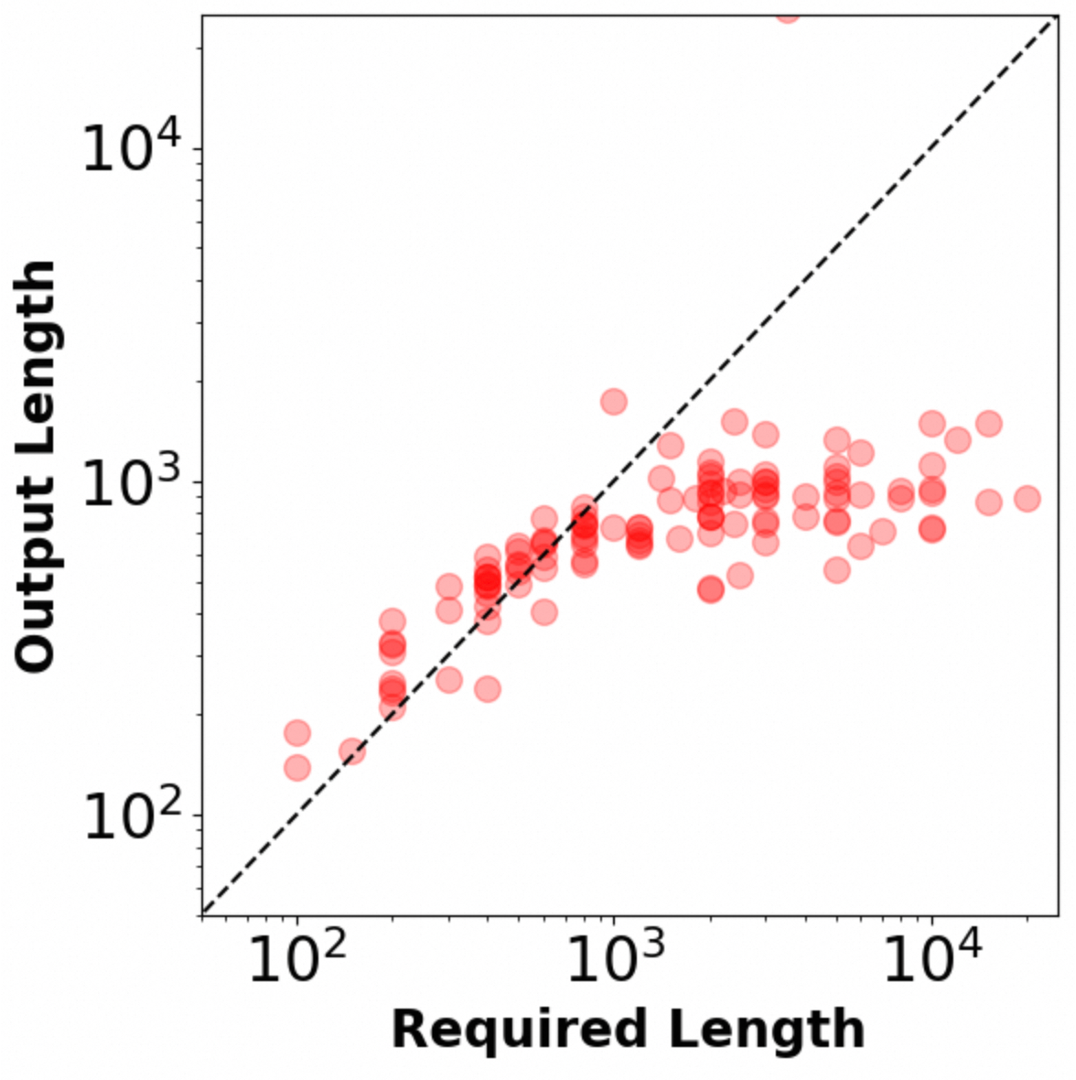}  
        \caption{Characteristics for \texttt{Qwen2-7B-Instruct}}
        \label{fig:qwen2-data-characteritics-sub1}
    \end{subfigure}
     \hfill  
    \begin{subfigure}[b]{0.23\textwidth}  
        \includegraphics[width=\textwidth]{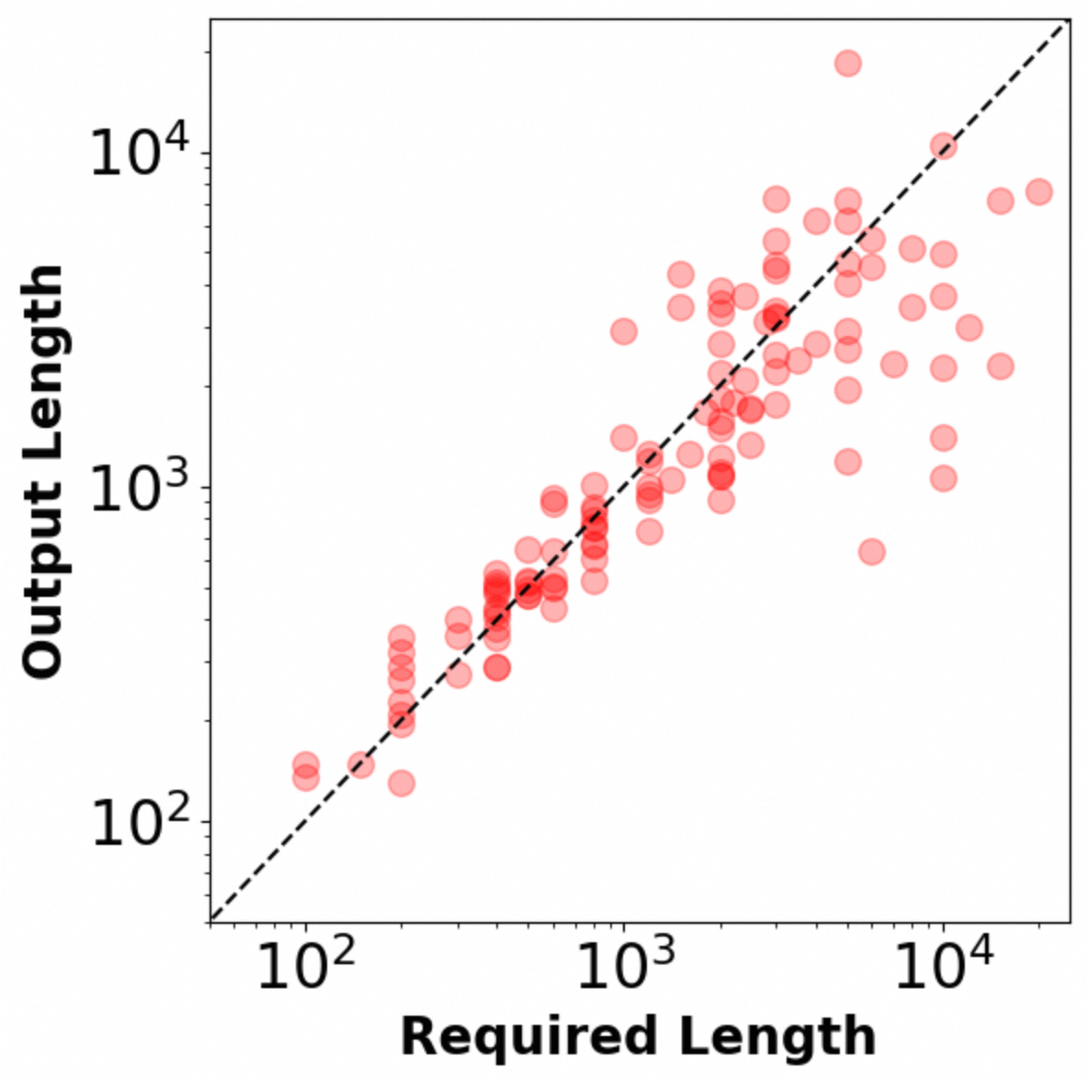}
        \caption{Characteristics for \texttt{MS-LongWriter-Qwen2-7B-Instruct}}
        \label{fig:qwen2-data-characteritics-sub2}
    \end{subfigure}
    \caption{\textmd{Improvement on ``length-following'' characteristics on \texttt{Qwen2-7B-Instruct}.}}
    \label{fig:qwen2-data-characteritics}
\end{figure}

Fig. \ref{fig:qwen2-data-characteritics} shows the length-following capability of the tuned model o f \texttt{MS-LongWriter-Qwen2-7B-Instruct} (from Setup 4 in Table \ref{tab:qwen2-result}) and the original \texttt{Qwen2-7B-Instruct} model. It can be observed that, the tuned model exhibits a  better adherence to prompt requirements on text output length. In particular, the model aligns much better to the length-instruction when the required text length is longer.

For a fair comparison between our approach summarized in Setup 4, and the original LongWriter model training\cite{bai2024longwriterunleashing10000word}, Table \ref{tab:glm4-result} shows how the  performance of the tuned compares. In particular, with \texttt{Setup} 4, we depart from the chat version of GLM4 model, \texttt{GLM4-9b-Chat}, and tuned the model with $666$ task data entries from \texttt{LongWriter-6k-filtered} and $13,320$ alignment data entries sampled from \cite{magpie-qwen2-chinese,magpie-qwen2-english}. Upon $2$ epochs of tuning, the training data instances amount to a total of $27,972$. The resulting model achieve a similar performance comparing to \texttt{LongWriter-GLM4-9B}, which was trained from GLM-4-9B with $6,000$ data samples from \texttt{LongWrite-6k}, and $180,000$ alignment data, for $4$ epochs, which amounts to a total of $744,000$ training data instances. In other words, similar performance is achieved here with a mere of \textbf{$\mathbf{3.74\%}$} of training instances, or in other words, about \textbf{$\mathbf{3.74\%}$ of compute cost} for training.  Clearly, the approach to train from a base model will necessitate larger amount of alignment dataset. However, with the result shown in Table \ref{tab:glm4-result}, we would like to argue that, for the specific task of long-form writing, a much cost-efficient tuning approach is possible by leveraging higher-quality tuning data, and by starting from a human-aligned version of model. The tuned model from \texttt{Setup} 4 of Table \ref{tab:glm4-result} is named \texttt{MS-LongWriter-GLM4-9B-Chat} and made publicly available at \cite{MS-LongWriter-GLM4-9B-Chat}.

Finally, we complete our experiments by applying our approach to the latest addition to the high-performing language model, the Qwen 2.5 family\cite{qwen2.5}. In particular, Table \ref{tab:qwen2.5-result} shows the result of applying our tuning recipes to \texttt{Qwen2.5-7B-Instruct}. Out of the box, \texttt{Qwen2.5-7B-Instruct} emerges as a strong competitor in all performance metrics under evaluation. This points to the perspective that, the evolution of general-purposed Instruct (or Chat) models can naturally lead to stronger capability in generating longer text. This coincide with our approach to tune long-writing capability based on human-aligned version of model, which is guided by the intuition that, the capability of writing long output is more complementary to,  rather than being in conflict with, the human-aligned capability of a model. Still, our tuning recipes demonstrates a notable performance improvement (for both \texttt{Setup} 4 and 5) against the stronger baseline of \texttt{Qwen2.5-7B-Instruct}. Specifically, on top of \texttt{Setup} 4 , we introduce an additional annealing phase with continuous-tuning in \texttt{Setup} 5. This results in a model that exhibits strongest performance in both $S_L$ and $S_Q$, we name this model \texttt{MS-LongWriter-Qwen2.5-7B-Instruct} and made it publicly available at \cite{MS-LongWriter-Qwen2.5-7B-Instruct}.

\section{Conclusion}
\label{sec:conclusion}
In this work, we explore efficient tuning of large language models (LLMs) to generate extended outputs that align with length-based instructions. Our findings highlight that high-quality datasets, specifically curated to match both length and coherence requirements, are essential for effective tuning. Notably, we demonstrate that significant performance gains can be achieved with minimal tuning effort, with high-quality dataset constitute of as few as $666$ data samples. We also demonstrate that similar performance improvement can be achieved with our approach using much less training data instances, comparing to the alignment-based approach. In this regard, our results suggests that starting from a well-trained, human-aligned model serves as an efficient foundation for unlocking long-form output capabilities. It is also worth investigating leveraging human-aligned models in tuning models targeting a broader range of other tasks.

\bibliographystyle{plain}
\bibliography{longwriter.bib}

\onecolumn
\appendix
\section{Appendix}
\subsection{Training Scripts}
\label{subsec:codes}
\subsubsection{Tuning with MS-Swift}
Example MS-Swift script for model tuning.
\begin{itemize}
    \item \texttt{Qwen2-7B-Instruct} model, and a $1:10:10$ mixing of the entire \texttt{LongWriter-6K-filtered} dataset (666 data entries), 6660 random data samples from \cite{magpie-qwen2-chinese} and 6660 random data samples from \cite{magpie-qwen2-english}. Refer to documentation of MS-Swift\cite{swift-doc} for detailed explanation of script parameters.
    \begin{lstlisting}[language=sh]
swift sft \
    --model_type qwen2-7b-instruct \
    --dataset longwriter-6k-filtered qwen2-pro-zh#6660 qwen2-pro-en#6660 \
    --max_length 28672 \
    --num_train_epochs 2 \
    --eval_steps 200 \
    --batch_size 1 \
    --gradient_accumulation_steps 64 \
    --gradient_checkpointing true \
    --warmup_ratio 0.1 \
    --learning_rate 1e-5 \
    --sft_type full \
    --loss_name long-ce \
    --check_dataset_strategy warning \
    --save_only_model false \
    --save_total_limit -1 \
    --lazy_tokenize true \
    --dataloader_num_workers 1 \
    --resume_only_model true \
    --neftune_noise_alpha 5 \
    --use_flash_attn true
\end{lstlisting}
    \item Continue from checkpoint out of Setup 4,  with additional annealing using new learning rate $2e-6$) for additional 2 epochs of \texttt{LongWriter-6k-filtered} dataset.
    \begin{lstlisting}[language=sh]
swift sft \
    --model_type qwen2_5-7b-instruct \
    --dataset longwriter-6k-filtered \
    --max_length 28672 \
    --num_train_epochs 2 \
    --eval_steps 200 \
    --batch_size 1 \
    --gradient_accumulation_steps 64 \
    --gradient_checkpointing true \
    --warmup_ratio 0.1 \
    --learning_rate 2e-6 \
    --sft_type full \
    --loss_name long-ce \
    --check_dataset_strategy warning \
    --save_only_model false \
    --save_total_limit -1 \
    --lazy_tokenize true \
    --dataloader_num_workers 1 \
    --resume_only_model true \
    --neftune_noise_alpha 5 \
    --use_flash_attn true \
    --resume_from_checkpoint {previous-checkpoint-path} > {output-checkpoint-path}
\end{lstlisting}
\end{itemize}

\subsection{Evaluation Script}
\label{subsec:evaluationscript}
    \begin{lstlisting}[language=python]

# Task Configuration
# `infer`--Inference; `eval_l`--S_L score evaluation; `eval_q`:S_Q score evaulation
task_cfg = dict(stage=['infer', 'eval_l', 'eval_q'],
                model='ZhipuAI/LongWriter-glm4-9b',
                input_data_path=None,
                output_dir='./outputs',
                openai_api_key=None,
                openai_gpt_model='gpt-4o-2024-05-13',
                infer_generation_kwargs={
                    'max_new_tokens': 32768,
                    'temperature': 0.5
                },
                eval_generation_kwargs={
                    'max_new_tokens': 1024,
                    'temperature': 0.5,
                    'stop': None
                },
                proc_num=8)

# execute evaluation task
from evalscope.third_party.longbench_write import run_task
run_task(task_cfg=task_cfg)
\end{lstlisting}

\end{document}